\documentclass[10pt,twocolumn,letterpaper]{article}

\usepackage{iccv}
\usepackage{times}
\usepackage{epsfig}
\usepackage{graphicx}

\usepackage{amsmath}
\usepackage{amsfonts}
\usepackage{enumitem}
\usepackage{mathrsfs}
\usepackage{bm}
\usepackage{cite}
\usepackage{authblk}

\usepackage{placeins}
\usepackage{multirow}
\usepackage{booktabs}
\DeclareMathOperator*{\argmaxB}{argmax} 

\usepackage[algo2e]{algorithm2e} 
\usepackage{algorithm}
\usepackage{algorithmic}

\usepackage{mathtools}

\SetKwProg{ParseTree}{ParseTree}{}{End}

\usepackage[pagebackref=true,breaklinks=true,colorlinks,bookmarks=false]{hyperref}
\iccvfinalcopy 

\def\iccvPaperID{8190} 

\ificcvfinal\fi

\begin{document}
\setlength{\abovedisplayskip}{2pt}
\setlength{\belowdisplayskip}{2pt}
\title{Auto-Parsing Network for Image Captioning and Visual Question Answering}
\author{Xu Yang$^{1,}$\thanks {Both authors contributed equally to this research.}  \hspace{1em} Chongyang Gao$^{ 2,*}$ \hspace{1em} Hanwang Zhang$^{3}$ \hspace{1em} Jianfei Cai$^{4}$ }
\affil{ \vspace{-0.5em} $^1$ School of Computer Science and Engineering, Southeast University
}
\affil{
\vspace{-0.5em} $^2$ Department of Computer Science, Dartmouth College}
\affil{
\vspace{-0.5em} $^3$ School of Computer Science and Engineering, Nanyang Technological University}
\affil{\vspace{-0.5em} $^4$ Department of Data Science and AI, Monash University}
\affil{
\vspace{-0.5em} \tt\small xuyangaca@gmail.com \hspace{0.2em} chongyang.gao.gr@dartmouth.edu \hspace{0.2em} hanwangzhang@ntu.edu.sg \hspace{0.2em} Jianfei.Cai@monash.edu}

\maketitle
\ificcvfinal\thispagestyle{empty}\fi

\begin{abstract}
   We propose an Auto-Parsing Network (APN) to discover and exploit the input data's hidden tree structures for improving the effectiveness of the Transformer-based vision-language systems. Specifically, we impose a Probabilistic Graphical Model (PGM) parameterized by the attention operations on each self-attention layer to incorporate sparse assumption. We use this PGM to softly segment an input sequence into a few clusters where each cluster can be treated as the parent of the inside entities. By stacking these PGM constrained self-attention layers, the clusters in a lower layer compose into a new sequence, and the PGM in a higher layer will further segment this sequence. Iteratively, a sparse tree can be implicitly parsed, and this tree's hierarchical knowledge is incorporated into the transformed embeddings, which can be used for solving the target vision-language tasks. Specifically, we showcase that our APN can strengthen Transformer based networks in two major vision-language tasks: Captioning and Visual Question Answering. Also, a PGM probability-based parsing algorithm is developed by which we can discover what the hidden structure of input is during the inference.
\end{abstract}

\section{Introduction}
\begin{figure}[t]
\centering
\includegraphics[width=1\linewidth,clip]{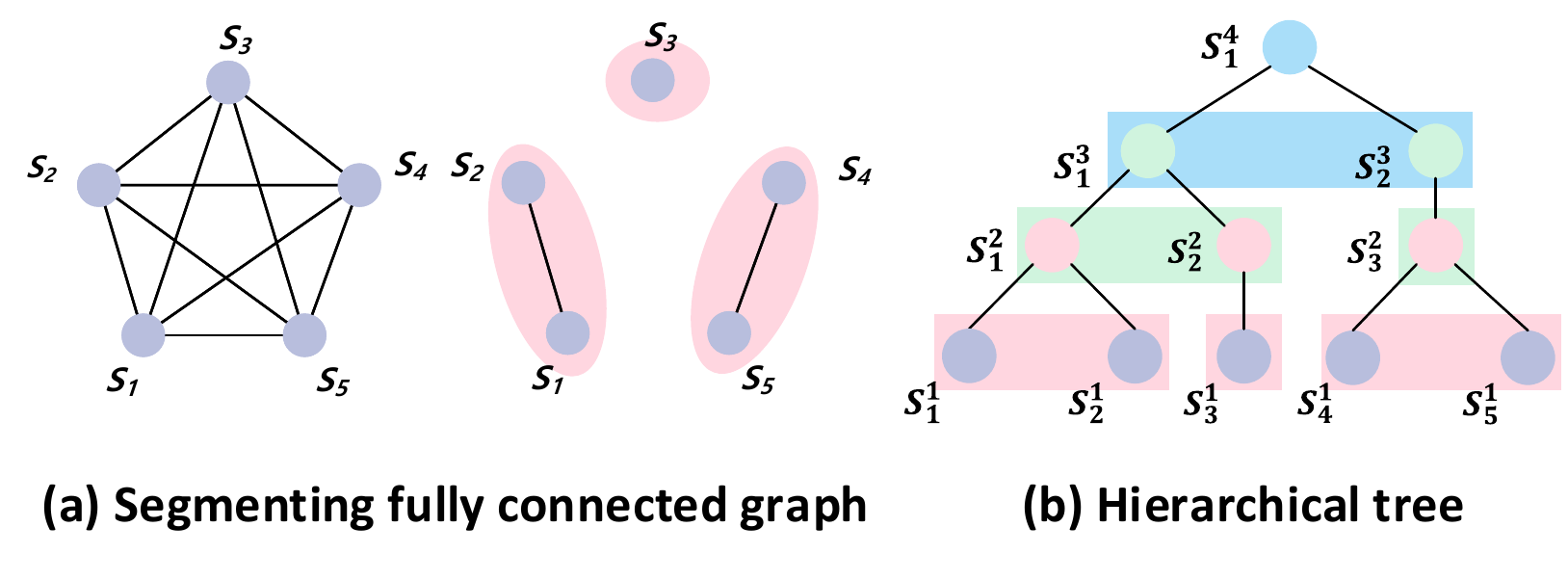}
   \caption{(a) The different graph priors entailed in the classic self-attention and our Probabilistic Graphical Model (PGM) constrained self-attention. Left: The classic self-attention pairs every two nodes in a graph and thus forms a fully connected graph. Right: Constrained by PGM, five nodes are segmented into three clusters. (b) By stacking our PGM constrained self-attention layers, a hierarchical tree can be automatically constructed. Hence we call our network as Auto-Parsing Network (APN).}
\label{fig:fig_segt}
\vspace{-0.2in}
\end{figure}
Nowadays, Transformer~\cite{vaswani2017attention} based frameworks have been prevalently applied into vision-language tasks and impressive improvements have been observed in image captioning~\cite{cornia2020meshed,pan2020x,herdade2019image,Li_2019_ICCV}, VQA~\cite{yu2019deep}, image grounding~\cite{ye2019cross,liu2019improving}, and visual reasoning~\cite{alberti2019fusion,tan2019lxmert}. Researchers attribute the progress to the various advantages of Transformer, like the efficient parallel computing~\cite{vaswani2017attention}, the ability to approximate any sequence-to-sequence function~\cite{yun2019transformers}, and the exploitation of the fully connected graph prior~\cite{battaglia2018relational} provided by self-attention as shown in the left part of Figure~\ref{fig:fig_segt}(a). Particularly, by the graph prior, although vision and language data have quite different superficial forms, their structural commonalities can be automatically abstracted, embedded, and transferred to narrow the domain gap.

However, the underlying structures of vision and language data are usually hierarchical and sparse, which are different from fully connected graphs, \eg, a sentence or an image can be parsed as constituent trees of words or objects, respectively~\cite{battaglia2018relational,chomsky2014aspects,tu2005image}. Without sparse and hierarchical constraints, this system may be overwhelmed by the trivial global dependencies and overlook the critical local context~\cite{li2019enhancing,xu2019leveraging,yang2019context}. Taking VQA as an example, the question in the last row of Figure~\ref{fig:fig_vis_ca} asks, ``What number is the hour hand on?'' for an image containing both hour hand and minute hand. A system with the fully connected graph prior may make an incorrect prediction by directly exploiting the global dependency between ``number'' and ``hand'' due to their high co-occurrence frequency in the training set, and thus neglects the key local context ``hour hand''. A similar problem is also observed in image captioning where noisy image scene graphs constructed by trivial dependencies may contribute less to the improvement~\cite{milewski2020scene}.

To reduce trivial connections of fully connected graphs, researchers usually parse the input into some sparse and hierarchical structures, such as the filtered scene graphs~\cite{johnson2015image, chen2020say,zhang2017visual} or sparse trees~\cite{tang2019learning} and then exploit them into solving various vision and language tasks, \eg, image captioning~\cite{yang2019auto,yao2018exploring}, VQA~\cite{andreas2016neural,li2019relation,teney2017graph}, grounding~\cite{bajaj2019g3raphground,liu2019learning} and VCR~\cite{yu2019heterogeneous}. However, these strategies require a large number of matched graph annotations~\cite{krishna2017visual,marcus-etal-1993-building,xue2005penn} for training useful parsers~\cite{zhu2013fast,chen2014fast,zellers2018neural,tang2020unbiased}; otherwise the domain shift can be induced, invalidating the parsed graphs.

To relieve the burden of incorporating hierarchical and sparse graph priors, inspired by Tree-Transformer~\cite{wang2019tree}, we propose a network that can learn to automatically parse inputs into trees during the end-to-end training without any additional graph annotations, hence named as \textbf{Auto-Parsing Network (APN)}. Specifically, we constrain the self-attention operation by a Probabilistic Graphical Model (PGM)~\cite{lafferty2001conditional,bishop2006pattern}, which is parameterized by differentiable attention operations. As shown in the right part of Figure~\ref{fig:fig_segt}(a), the PGM helps to segment the input sequence into a few clusters. After each segmenting iteration, only the entities in the same cluster can attend to each other, and thus, the local context is embedded. Intuitively, each cluster can be considered as the parent of inside entities and these clusters together compose a new sequence. By stacking constrained self-attention layers, the new sequence at a lower layer will be further segmented by the PGM at a higher layer. For example, as in Figure~\ref{fig:fig_segt} (b), $s_1^1$ and $s_2^1$ in the first level are clustered into a new pseudo-parent node $s_1^2$ in the second level. Then, $s_1^2$ and $s_2^2$ are further clustered. Via this iterative way, a tree can be automatically parsed. 

By APN, the local and global contexts can be accordingly embedded at lower and higher layers. Once we build an encoder-decoder based on APN, both source and target domains' hierarchical structures can be automatically parsed, embedded, and transferred. We deploy the proposed APN in two fundamental vision-language tasks: image captioning~\cite{vinyals2015show, xu2015show}, and visual question answering~\cite{antol2015vqa}. Experiment results on both tasks show that our APN obtains consistent improvements compared to Transformer based models. Furthermore, we develop a parsing algorithm, which can generate constituent trees for vision and language inputs based on the calculated PGM probabilities. In this way, when the model infers, the hidden structure for each sample can be revealed.

In summary, we have the following contributions:
\begin{itemize}[leftmargin=.1in]
\item Inspired by Tree-Transformer~\cite{wang2019tree}, we propose an Auto-Parsing Network (APN) which can unsupervisedly learns to parse trees for the inputs by imposing PGM probabilities on self-attention layers and exploiting hierarchical constraints into PGM probabilities.
\item We design two different APNs for solving Image Captioning and visual Question Answering.
\item We show that our APN achieves consistent improvements compared with the classic Transformer on both tasks.
\end{itemize}

\section{Related Work}
\noindent\textbf{Structured Learning.}
Structured learning is a research hotspot~\cite{nowozin2011structured,kulesza2008structured} since most data have hidden structures and the exploitation of these structures is beneficial to solving downstream tasks~\cite{botvinick2008hierarchical,tenenbaum2011grow}.
The Probabilistic Graphical Model is one classic method which has been incorporated into deep networks to solve vision~\cite{chen2014semantic,koller2009probabilistic} and NLP tasks~\cite{huang2015bidirectional,wang2016towards}. However, some of such models need to use the forward-backward algorithm~\cite{stratonovich1965conditional} or the inside-outside algorithm~\cite{baker1979trainable} for inferring the probability during the training, which will unavoidably slow the training. 

To accelerate the training, we follow \textbf{Tree-Transformer}~\cite{wang2019tree} which applies the attention mechanism to calculate a constraint matrix to achieve soft segmentation. However, their constraint matrix is not a normalized probability and we reformulate their technique into a form of PGM. Based on this PGM probability, we design two different Auto-Parsing Networks to respectively solve image captioning and visual question answering. In this way, we can implicitly parse the hidden trees from the input data and the networks can be trained end-to-end without using the forward-backward or inside-outside algorithms.

\noindent\textbf{Exploiting Graphs in Visual Reasoning.}
Image Captioning~\cite{vinyals2015show, xu2015show} and Visual Question Answering~\cite{antol2015vqa} are two fundamental tasks in visual reasoning, that aim to generate a fluent sentence to describe a visual scene and provide answers to questions related to the visual contents, respectively. Thanks to the blooming of deep learning, encoder-decoder structures~\cite{yang2019auto}, attention mechanisms~\cite{anderson2018bottom} including self-attention~\cite{huang2019attention} and many other techniques, the performance of image captioning and VQA has been boosted significantly.  Furthermore, considering that graph priors can transfer commonalities and mitigate the gap between visual and language domains, researchers explore how to use graphs~\cite{teney2017graph,yu2020ernie} properly in both tasks. On one hand, graphs can be transferred into latent variables by GCN~\cite{yao2018exploring,yang2019auto}, which can be directly utilized by models. On the other hand, the input images and questions can be parsed into trees~\cite{wang2019tree}. Because of structural correspondence, deep module networks~\cite{yu2019deep, hu2017learning} can feed different types of parts of a parse tree into separate modules. However, the needs for additional data and parsers limit the use of graph priors. To address this limitation, we parse inputs into trees based on PGMs and self-attention for different tasks automatically, without the need for any extra data and parsers.

\section{Multi-Head Attention Revisit}

We first revisit the multi-head attention operation, which is the elemental building block of our APN. Given the query, key, and value matrices, $\bm{Q} \in \mathbb{R}^{N_Q \times d}$, $\bm{K} \in \mathbb{R}^{N_K \times d}$, $\bm{V} \in \mathbb{R}^{N_K \times d}$,\footnote{The number of the elements in $\bm{K}$ and $\bm{V}$ are the same and we use $N_K$ to denote it.} Multi-Head attention~\cite{vaswani2017attention} calculates an attended matrix $\bm{O}$ as follows:
\begin{equation} \label{equ:multi-head}
\small
\begin{aligned}
 \textbf{Input:} \quad  &\bm{Q},\bm{K},\bm{V} \\
 \textbf{Att:} \quad  &\bm{A}_i=\text{Softmax}( \frac{\bm{Q}\bm{W}_i^Q(\bm{K}\bm{W}_i^K)^T}{\sqrt{d}} ) \\
 \textbf{Head}:  \quad  &\bm{H}_i=\bm{A}_i\bm{V}\bm{W}_i^V,\\
 \textbf{Multi-Head:} \quad & \bm{H}= [\bm{H}_1,\bm{H}_2,...,\bm{H}_8]\bm{W}^H, \\
 \textbf{Output:} \quad  &\bm{O}=\text{FFN}(\bm{H}), \\
\end{aligned}
\end{equation}
where $\bm{W}_i^Q, \bm{W}_i^K, \bm{W}_i^V \in \mathbb{R}^{d \times d_h}$, and $\bm{W}_i^H \in \mathbb{R}^{  d \times d}$ are all trainable matrices; the number of the heads is set to 8 and $d_h=d/8$; $\bm{A}_i$ is the attention matrix for calculating the $i$-th head $\bm{H}_i$; $[\cdot]$ is the concatenation operation; and FFN is the Position-wise Feed-Forward Network: FC-RELU-FC.

Specifically, when setting $\bm{Q}$, $\bm{K}$, $\bm{V}$ to the same value, such mechanism is called self-attention or non-local convolution~\cite{wiegreffe2019attention,wang2018non}. From the Graph Network perspective, it is building a fully-connected graph~\cite{battaglia2018relational}. For example, given a representation sequence $\bm{S}=\{\bm{s}_1,\bm{s}_2,...,\bm{s}_T\} \in \mathbb{R}^{d \times T}$ and by setting $\bm{Q}=\bm{K}=\bm{V}=\bm{S}$, each two nodes $\bm{s}_i$ and $\bm{s}_j$ are connected by an edge weighted by the attention weight $\alpha_{i,j}$. In this way, dense and long-term dependencies between each two entities are embedded, which has been proven to be beneficial in various tasks.

However, since the underlying structures of data are usually hierarchical and sparse, a fully connected graph assumption may lead to trivial attention patterns~\cite{vig2019visualizing}, \eg, only one entity is attended by the other entities, or the critical local context as discussed in Introduction is overlooked. To induce the sparse and hierarchical prior for capturing more meaningful relations, we impose the probabilistic graphical model (PGM) into the self-attention network and stack them to build an Auto-Parsing Network (APN).

\section{Probabilistic Graphical Model}

\begin{figure}[t]
\centering
\includegraphics[width=1\linewidth]{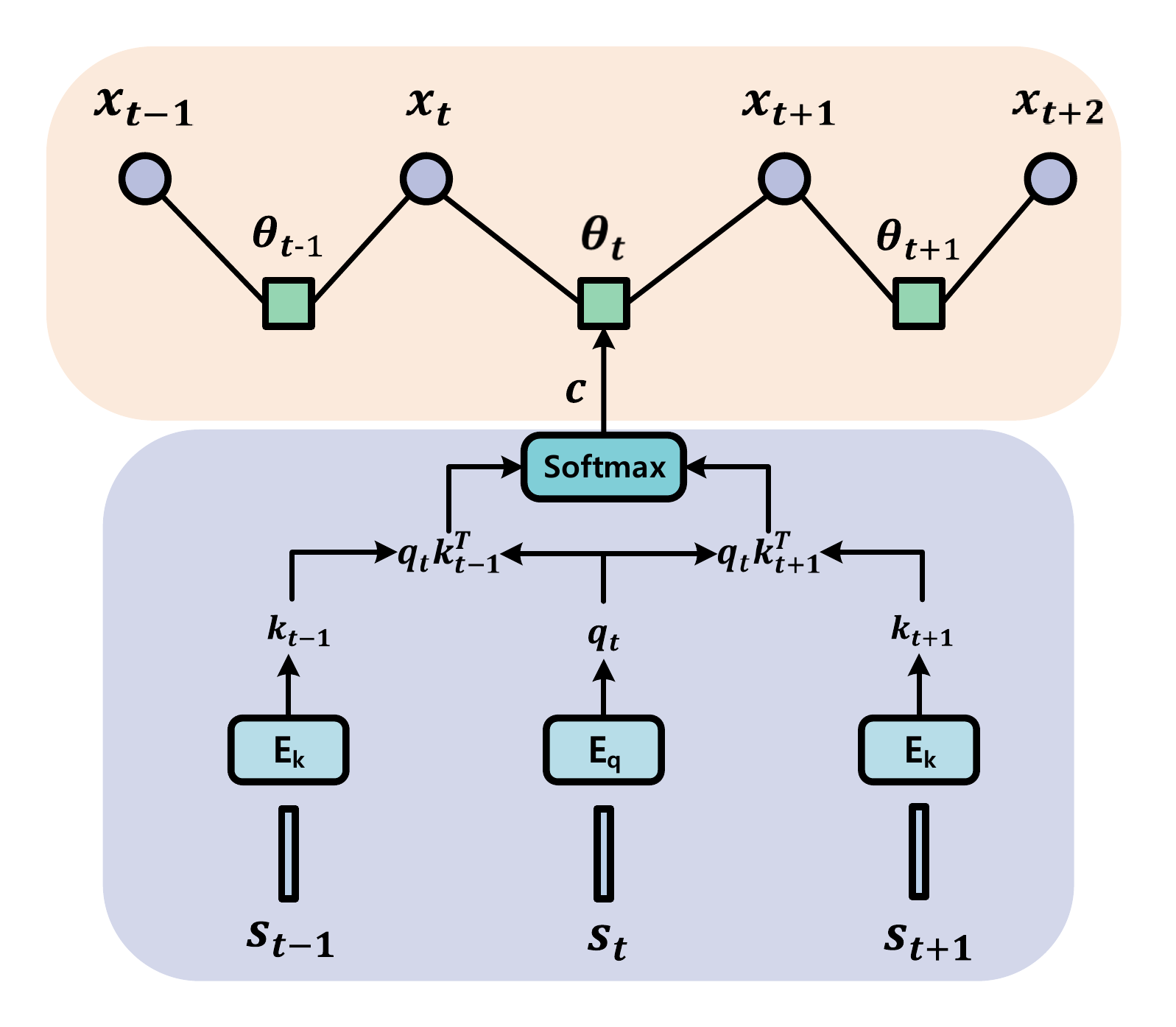}
   \caption{Illustration of our probabilistic graphical model (PGM), the top part is the PGM, and the bottom part is the implementation of the potential function. The \textbf{$E_k$} and \textbf{$E_q$} encode $\bm{s}$ into $\bm{k}$ and $\bm{q}$, respectively.}
\label{fig:fig_CR}
\end{figure}

Generally, a PGM is defined as the product of the clique potential functions that are all conditioned on an input set~\cite{lafferty2001conditional,murphy2012machine}:
\begin{equation} \label{equ:general_pgm}
    P(\bm{x}|\bm{S})=\frac{1}{Z(\bm{x},\bm{S})}\prod_{C}\theta_{C}(\bm{x}_C|\bm{S}),
\end{equation}
where $\bm{S}$ is the set of a sequential entities, $\theta_{C}(\cdot)$ is the potential function, and $Z(\cdot)$ is the partition function:
\begin{equation} \label{equ:partition_function}
    Z(\bm{x},\bm{S}) = \sum_{\bm{x}}\prod_{C}\theta_{C}(\bm{x}_C|\bm{S})
\end{equation}
which makes the overall probabilities sum to 1. 

In our case, since we focus more on the correlation between neighboring entities (\eg, whether a few neighboring entities should be grouped together), we apply the pairwise PGM where each clique only includes the neighboring two entities and the PGM formula can be simplified as:
\begin{equation} \label{equ:linear_pgm}
    P(x_1,x_2,..,x_{T-1}|\bm{S})=\frac{1}{Z(\bm{x},\bm{S})}\prod_{t=1}^{T-1}\theta_{t}(x_t|\bm{S}),
\end{equation}
where $x_t$ is a binary hidden variable denoting whether $\bm{s}_t$ and $\bm{s}_{t+1}$ is connected to each other and $\theta_t(\cdot)$ measures the correlation between $\bm{s}_t$ and $\bm{s}_{t+1}$, as visualized in the top part of Figure~\ref{fig:fig_CR}. For example, a large $\theta_3(x_3=1)$ indicates that $\bm{s}_3$ and $\bm{s}_4$ should be grouped together. 


\subsection{Network Parameterization of $\theta_{t}(\cdot)$}
Specifically, we set $\theta_{t}(\cdot)$ as:
\begin{equation}  \label{equ:bernoulli_distribution}
\small
\begin{aligned}
    \theta_{t}(x_t=1|\bm{S}) &=\text{Ber}(z_t^{t+1}=1|c_t^{t+1})\text{Ber}(z_{t+1}^{t}=1|c_{t+1}^t) \\
    \theta_{t}(x_t=0|\bm{S})&=1-\theta_{t}(x_t=1|\bm{S}),
\end{aligned}
\end{equation}
where $\theta_{t}(x_t=1|\bm{S})$ is the product of two Bernoulli distributions and $\theta_{t}(x_t=0|\bm{S})$ is set to guarantee that the sum is 1. In this way, each $\theta_{t}(\cdot)$ is normalized and ${Z(\bm{x},\bm{S})}$ in Eq.~\eqref{equ:linear_pgm} is always 1. The Bernoulli distribution $\text{Ber}(z|c)=c^z(1-c)^{(1-z)}$ and $z_t^{t+1}$ is a binary variable indicating whether $\bm{s}_t$ should be connected with $\bm{s}_{t+1}$ with the corresponding probability $c_t^{t+1}$. Eq.~\eqref{equ:bernoulli_distribution} is designed based on the following intuition: if an entity $\bm{s}_t$ is close to its right neighbor $\bm{s}_{t+1}$ (i.e. $z_t^{t+1}=1$) and that neighbor is also close to this entity (i.e. $z_{t+1}^{t}=1$), then they should be tightly connected ($x_t=1$).  

To determine whether an entity $\bm{s}_t$ is close to its right neighbor $\bm{s}_{t+1}$, we can compare their closeness with that between $\bm{s}_t$ and its left neighbor $\bm{s}_{t-1}$, which naturally derives the following attention network parameterization of the probability $c$:
\begin{equation} \label{equ:parameterization_c}
[c_t^{t-1},c_t^{t+1}] = \text{ATT}(\bm{q}_t,[\bm{k}_{t-1},\bm{k}_{t+1}]),
\end{equation}
where $\bm{q}_t = \bm{s}_t\bm{W}^q, \bm{k}_{t} = \bm{s}_{t}\bm{W}^k$. Inputting the sequential set $\bm{S}$ into this attention network, we can calculate $c_t^{t-1}$ and $c_t^{t+1}$ for each $\bm{s}_t$. As in the bottom part of Figure~\ref{fig:fig_CR}, by bringing these probabilities back into Eq.~\eqref{equ:bernoulli_distribution} and~\eqref{equ:linear_pgm}, we can calculate the values of the potential functions and the $P(\bm{x}|\bm{S})$ for any binary hidden vector $\bm{x}$.

\subsection{Segmenting a Sequence}
An intuitive strategy of using the PGM with Eq.~\eqref{equ:linear_pgm} to segment a sequence is to find the maximum of this probabilistic model:
\begin{equation} \label{equ:find_mode}
    \bm{x}^*=\argmaxB_{\bm{x}}(P(\bm{x}|\bm{S})).
\end{equation}
However, solving Eq.~\eqref{equ:find_mode} over all possible $x$ using traditional methods like max-sum algorithm~\cite{bishop2006pattern} is not fast and moreover it needs to be repeated in each forward pass during training, which is unbearable.

Instead, similar to the technique proposed in tree-transformer~\cite{wang2019tree}, we approximate this hard segmentation in a soft manner. Specifically, we calculate a $T\times T$ matrix $\bm{M}$ where if $i<j$:
\begin{equation} \label{equ:mij}
\begin{aligned}
M_{i,j}&=P(x_i=1,x_{i+1}=1,...,x_{j-1}=1) \\
&=\prod_{t=i}^{j-1}\theta_{t}(x_t=1|\bm{S}), \\
\end{aligned}
\end{equation}
if $i>j$, $M_{i,j}=M_{j,i}$, and $M_{i,i}=1$. In this matrix, each $M_{i,j}$ is the marginal probability of clustering the entities from $\bm{s}_i$ to $\bm{s}_j$, \eg, $M_{2,4}$ measures the probability of clustering the entities $\bm{s}_2,\bm{s}_3,\bm{s}_4$, and then if $M_{2,4}$ is large, $\bm{s}_2,\bm{s}_3,\bm{s}_4$ are softly clustered.

After calculating $\bm{M}$, we use it to revise the Head operation in Eq.~\eqref{equ:multi-head} as follows:
\begin{equation}
    \bm{H}=(\bm{M} \otimes \bm{A})\bm{V}\bm{W}^V,
\end{equation}
where $\otimes$ stands for the element-wise product. Intuitively, after revising, the original fully-connected graph becomes a sparser graph that contains a few clusters and the entities are only connected with each other in the same cluster. For example, as in the left of Figure.~\ref{fig:fig_segt} (a), $\bm{s}_i$ and $\bm{s}_j$ can be freely attended by each other in self-attention where the attended weights are $A_{i,j}$ and $A_{j,i}$, while in the right part, after constraining by $\bm{M}$, they can only attend each other if $M_{i,j}$ has a large value, which denotes $\bm{s}_i$ and $\bm{s}_j$ are softly clustered, \eg, $\bm{s}_1$ and $\bm{s}_2$ belong to the same cluster. In this way, the whole sequence is softly segmented into a few clusters. Note that we use the same $\bm{M}$ to constrain the self-attention weights of different heads since our PGM aims to segment the input sequence and all the heads should follow the segmented clusters to calculate the outputs.

\subsection{Parsing a Tree}
In this section, we first introduce how to automatically and implicitly parse a tree during the forward pass. Then we introduce an algorithm for explicitly parsing a tree after the forward pass, which can visualize how the hidden tree structure is incorporated into the embeddings.

\noindent\textbf{Implicitly Parsing during Forward Pass.}
We have shown that by multiplying the self-attention weights with our PGM probability matrix $\bm{M}$, the input sequence can be softly segmented into clusters. The segmented clusters can be considered as the parents of the entities in it, \eg, as shown in Figure~\ref{fig:fig_segt} (b), $\bm{s}_1^{2}=\{\bm{s}_1^{1},\bm{s}_2^{1}\}$ denotes that the parent $\bm{s}_1^{2}$ has two children: $\bm{s}_1^{1},\bm{s}_2^{1}$. 

However, simply stacking our PGM constrained layers cannot parse a tree since the entities in a lower-layer cluster may not still be in a  higher-layer cluster, \eg, $M_{i,j}^{l}$ is large while $M_{i,j}^{l+1}$ is small. To amend this drawback, following Tree-Transformer~\cite{wang2019tree}, we modify the potential function  (Eq.~\eqref{equ:bernoulli_distribution}) at the $l$-th level as:
\begin{equation} \label{equ:convex_combination}
    \tilde{\theta}_t^l =  \tilde{\theta}_t^{l-1} + (1- \tilde{\theta}_t^{l-1})\theta_t^{l},
\end{equation}
where $\theta_t^l$ is the abbreviation of $\theta_t^l(x_t^l=1|\bm{S}^l)$. Since $0 \leq \tilde{\theta}_t^{l-1} \leq 1$, $\tilde{\theta}_t^l$ is the convex combination of 1 and $\theta_t^l$, $\tilde{\theta}_t^l$ will always be larger or equal than $\theta_t^l$. Then we replace $\theta_t^l$ with $\tilde{\theta}_t^l$ in Eq.~\eqref{equ:mij} to compute $\bm{M}^{l}$ and will have $M_{i,j}^{l+1} \geq M_{i,j}^{l}$. Then the entities which are clustered at a lower layer will still be clustered at a higher layer. By stacking these constrained layers, during the iterative segmenting, an implicitly tree can be automatically parsed during the forward pass. One toy example about this is shown in Figure~\ref{fig:fig_segt} (b).

\noindent\textbf{Explicitly Parsing after Forward Pass.}
With an inputting sequence, APN not only outputs the embeddings that incorporate the hidden hierarchical knowledge through the forward pass, but also calculates a series of potential probabilities~(Eq.~\eqref{equ:convex_combination}) in each layer. We follow~\cite{wang2019tree} to provide an algorithm to explicitly parse a sequence into a tree by potential probabilities, which is given in Algorithm~\ref{alg1}. This algorithm parses a sequence into a tree from top to bottom recursively. $\textbf{ParseTree}(l, i, j)$ denotes the process of using the potential function of the $l$-th level  $\tilde{\theta}^l$ to segment a sequence that begins at $\bm{s}_i$ and ends at $\bm{s}_j$. In line 1, it first determines whether $\bm{s}_i$ and $\bm{s}_j$ are neighbors: if they are, it segments $\bm{s}_i$ and $\bm{s}_j$ as the left and right leaves, respectively; otherwise, in Line 4, it finds the position $p^{*}$ that has the minimum potential function value to segment the sequence into two parts: $\{\bm{s}_{i:p^{*}}\}$ and $\{\bm{s}_{p^{*}+1:j}\}$, as in Line 12 and 13. Figure~\ref{fig:fig_vis_ca} shows the parsed results for captioning and VQA.

\begin{algorithm}[tb]

 \caption{Parsing Tree by Potential values}
 \label{alg1}
 \hspace*{0.02in}{\bf Input:}  $\tilde{\theta}_t^l$ for $t=1:T,l=1:L$\\
$id$ $\leftarrow$  minimum layer id \\
$threshold$ $\leftarrow$ Split point's threshold\\
$l$ $\leftarrow$ layer index; $i$ $\leftarrow$ left position; $j$ $\leftarrow$ right position\\
 \hspace*{0.02in}{\bf Output:} 
 ParseTree \\
 \hspace*{0.02in}{\bf ParseTree$(l, i, j):$} 
 \begin{algorithmic}[1]
  
  \IF{$i-j \leq 1$}
  \STATE {\textbf{return} $(i, j)$}
  \ENDIF
  \STATE {$p^* \leftarrow \textbf{argmin}_{\bm{p=i,...,j-1}}(\tilde{\theta}_{p}^l)$}
  \STATE {$next \leftarrow \textbf{max}(l-1, id)$}\\
  \IF{$\tilde{\theta}_{p^*}^l > threshold$}
  \IF{$l == id$}
  \STATE {\textbf{return} $(i, j)$}
  \ENDIF
  \STATE {\textbf{return ParseTree}$(next, i, j)$}
  \ENDIF
  \STATE {$LeftTree \leftarrow \textbf{ParseTree}(next, i, p^*)$}
  \STATE {$RightTree \leftarrow \textbf{ParseTree}(next, p^* + 1, j)$}
  \STATE {\textbf{return} $(LeftTree, RightTree)$}

 \end{algorithmic}
\end{algorithm}

\section{Image Captioning}
\label{sec:ImgCap}
\subsection{Architecture and Objectives}

\begin{figure}[t]
\vspace{-0.15in}
\centering
\includegraphics[width=1\linewidth,trim = 5mm 3mm 5mm 5mm,clip]{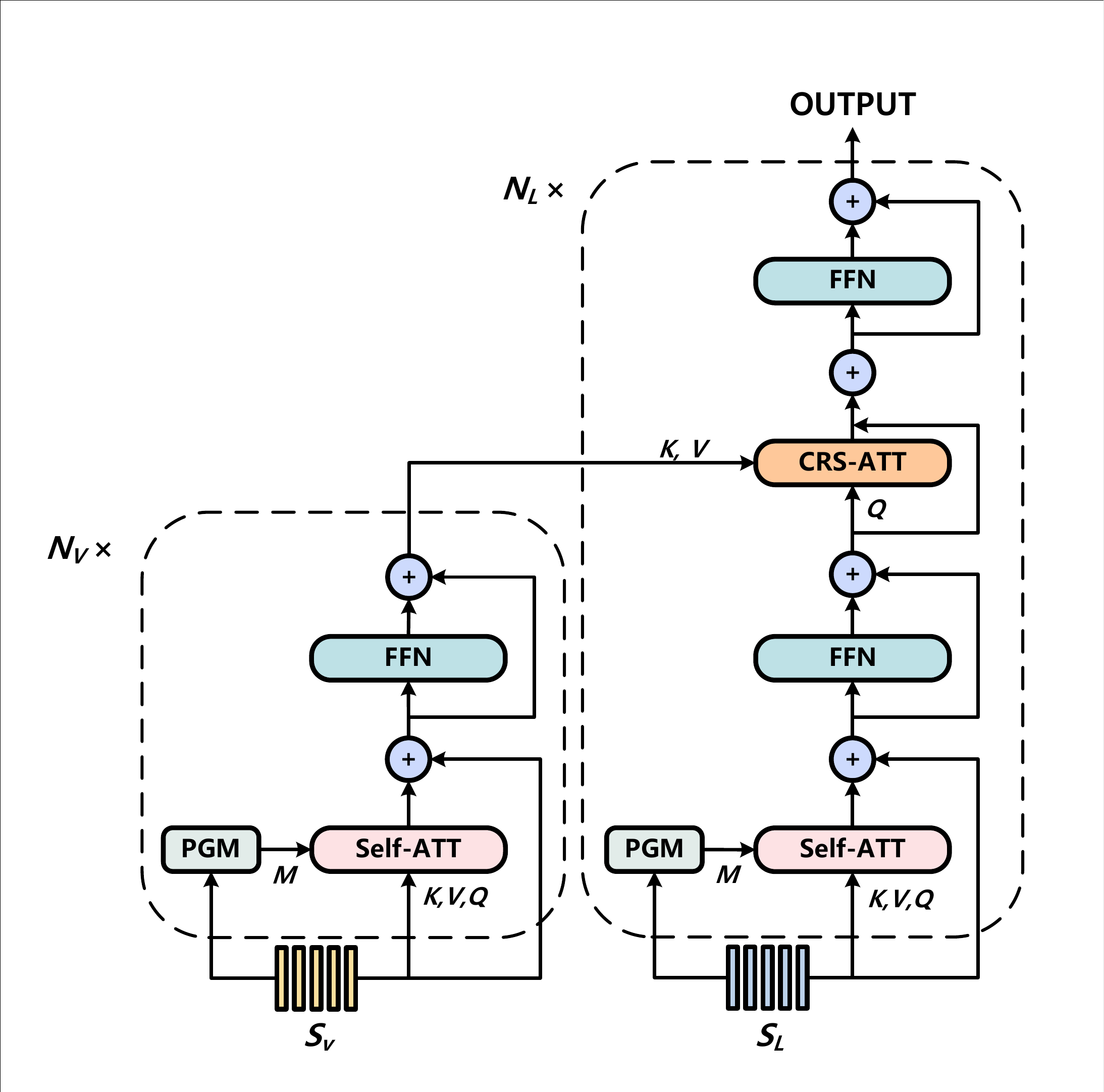}
   \caption{An overview of our APN for image captioning task. The FFN, CRS-ATT and Self-ATT denote the feed-forward network, cross-attention network and self-attention network. }
\label{fig:fig_capach}
\vspace{-0.15in}
\end{figure}

The APN architecture for image captioning is shown in Figure~\ref{fig:fig_capach}. $S_V$ and $S_L$ are respectively the visual and language representation sequences. Note that the $S_V$ is linearized by sorting the RoIs from top-left to bottom-right. The left and right parts sketch the visual encoder and the language decoder, which are both stacked by 6 blocks, $N_V=N_L=6$. We first train our APN by minimizing the cross-entropy loss:
\begin{equation}
    L_{CE} = -\log P({Y}^*),
\label{equ:equ_celoss}
\end{equation}
where ${Y}^*$ is the ground truth caption for the given image. Then, the model is further trained by maximizing a reinforcement learning (RL) based reward~\cite{rennie2017self} :
\begin{equation}
    R_{RL} = \mathbb{E}_{Y^s \sim P(Y)}[r({Y}^s;{Y}^*)],
    \label{equ:equ_rlloss}
\end{equation}
where ${Y}^s$ is the sampled sentence and $r$ is a sentence-level metric, \eg, the CIDEr-D~\cite{vedantam2015cider} metric, for ${Y}^s$ and ${Y}^*$.

\subsection{Dataset, Settings and Metrics}
\noindent\textbf{Microsoft COCO Dataset}~\cite{lin2014microsoft} includes 123,287 images and each of them has 5 captions as labels. We conduct experiments on Karpathy split (113,287/5,000/5,000 train/val/test images) for offline testing and Official online test split (82,783/40,504/40,775 train/val/test images).

\noindent\textbf{Settings}. 
We preprocess the captions by the following steps. We change all words to lowercase and delete the words appearing less than 5 times. Then we trim the sentences to a maximum of 16 words. Finally, we get a vocabulary of $10, 369$ words. We use the visual features extracted by Up-Down~\cite{anderson2018bottom}. The dimensionalities for both encoder and decoder are $d_{h} = 512$ (Eq.~\eqref{equ:multi-head}); and the dimension of the inner-layer in feed-forward networks is $2048$. We multiply all heads' attention matrix by the same $M$ (Eq.~\eqref{equ:multi-head}).We use Adam~\cite{kingma2014adam} optimizer following the settings in~\cite{vaswani2017attention} with $warmup\_steps = 20,000$. We first use cross-entropy loss (Eq.~\eqref{equ:equ_celoss}) for 15 epochs and initialize the learning rate as $1e^{-5}$ and decay it by 0.8 every 5 epochs. Then the RL-based reward (Eq.~\eqref{equ:equ_rlloss}) is used for another 35 epochs where the learning rate is reset to $1e^{-5}$ and decay by 0.8 every 5 epochs. The batch size is 10.

\subsection{Results}

\noindent\textbf{Ablation Studies}.
We conduct ablation studies to validate the effectiveness of our two key components: the probabilistic graphical model probability and the hierarchical constraint. Specifically, we design and compare the following ablative models.
\textbf{BASE}: We use the architecture illustrated in Figure~\ref{fig:fig_capach} without PGM. \textbf{PGM}: We incorporate all the PGM modules into the BASE architecture while we do not use the hierarchical constraints in Eq.~\eqref{equ:convex_combination}. \textbf{APN}: We use the whole APN architecture where the potential probabilities are calculated by Eq.~\eqref{equ:convex_combination}.  

\noindent\textbf{Results and Analysis}.
To evaluate the quality of the generated image captions, we measure the similarities between the generated caption with the ground-truth captions by five metrics: CIDEr-D~\cite{vedantam2015cider}, BLEU~\cite{papineni2002bleu}, METEOR\cite{banerjee2005meteor}, ROUGE~\cite{lin2004rouge} and SPICE~\cite{anderson2016spice}. 

The bottom section of Table~\ref{table:tab_cap} shows the performances of various baselines on MS-COCO Karpathy split. Compared with BASE, our PGM has a better performance on nearly all metrics.  This suggests that the incorporated sparse assumption of our PGM can improve the qualities of the generated captions. APN achieves the best performance among all baselines, which demonstrates that the designed hierarchical property can constrain the attention to improve the model by enforcing a tree structure. 


To validate whether more local contexts are exploited than the self-attention based Transformer, we also evaluate the recalls of different kinds of words: objects, attributes, relations, and genders. Specifically, we calculate the recall by counting whether a word in ground-truth captions appears in the generated captions. Note that since an image is assigned with five captions and different captions may use distinctive words, while we measure the recall of these words in only one generated caption. Thus the more distinctive a kind of words, \eg, attribute, the recall is lower. We also exploit the bias measurements CHAIR~\cite{objectHallucination} to validate whether our APN can avoid trivial or even negative global dependencies coming from the dataset bias. 

All these scores are listed in Table~\ref{table:tab_ab_cap} and we can find that our APN has the highest recalls of each kind of word and the lowest CHAIR scores. Both results demonstrate that our APN can exploit the local context to generate more descriptive words, and meantime avoid the negative global dependencies to generate less biased captions.

\begin{table}[t]
\begin{center}
\caption{The CHAIR scores and captions' recall scores in terms of Object, Attribute, Relation and Gender of ablative models on MS-COCO Karpathy split. $\uparrow$ and $\downarrow$ mean the higher the better and the lower the better, respective.}
\label{table:tab_ab_cap}
\scalebox{0.65}{ 
\begin{tabular}{l c  c c c c c}
		\hline
           Models   & Object$\uparrow$ & Attribute$\uparrow$ & Relation$\uparrow$ & Gender$\uparrow$ & CHAIRs $\downarrow$ & CHAIRi$\downarrow$\\ \hline
           BASE & $28.2$ & $9.2$ & $22.3$ & $62.1$  & $12.4$  & $9.5$\\
           PGM     & $29.4$ & $10.3$ & $23.4$ & $63.3$ & $11.6$  & $8.7$\\
           APN     & $\mathbf{30.9}$ & $\mathbf{11.0}$ & $\mathbf{24.7}$ & $\mathbf{64.5}$ & $\mathbf{10.6}$  & $\mathbf{6.8}$\\ \hline
\end{tabular}
}
\end{center}
\vspace{-0.2in}
\end{table}

\begin{table}[t]
\begin{center}
\caption{The performances of various methods on MS-COCO Karpathy split. The metrics: B@N, M, R, C and S denote BLEU@N, METEOR, ROUGE-L, CIDEr-D and SPICE.}
\label{table:tab_cap}
\scalebox{0.8}{
\begin{tabular}{l  c c c c c}
		\hline
		   Models   & B@4 & M & R &   C & S\\ \hline
           GCN-LSTM~\cite{yao2018exploring}  & $38.2$ & $28.5$ & $58.3$ & $127.6$ & $22.0$ \\ 
           SGAE~\cite{yang2019auto}   & $38.4$ & $28.4$ & $58.6$ & $127.8$ & $22.1$ \\
           CNM~\cite{yang2019learning}   & $38.9$ & $28.4$ & $58.8$ & $127.9$ & $22.0$ \\
           HIP~\cite{yao2019hierarchy}   & $39.1$ & $28.9$ & $\mathbf{59.2}$ & $130.6$ & $22.3$ \\
           ETA~\cite{li2019entangled}    & $\mathbf{39.9}$ & $28.9$ & $58.9$ & $126.6$ & $22.7$ \\ 
           ORT~\cite{herdade2019image}   & $38.6$ & $28.7$ & $58.4$ & $128.3$ & $22.6$ \\
           AoANet~\cite{huang2019attention}   & $38.9$ & $\mathbf{29.2}$ & $58.8$ & $129.8$ & $22.4$ \\
           $\mathcal{M}^2$ Transformer~\cite{cornia2020meshed}    & $39.1$ & $\mathbf{29.2}$ & $58.6$ & $131.2$ & $22.6$ \\ 
           \hline
           BASE & $38.4$ & $28.5$ & $58.1$ & $128.7$ & $22.0$ \\
           PGM    & $38.9$ & $28.9$ & $58.5$ & $130.4$ & $22.6$\\
           APN   & $39.6$ & $\mathbf{29.2}$ & $59.1$ & $\mathbf{131.8}$ & $\mathbf{23.0}$\\ \hline

\end{tabular}
}
\end{center}
\vspace{-0.3in}
\end{table}

\begin{figure*}[t]
\centering
\includegraphics[height=12cm, width=1\linewidth,trim = 5mm 5mm 5mm 5mm,clip]{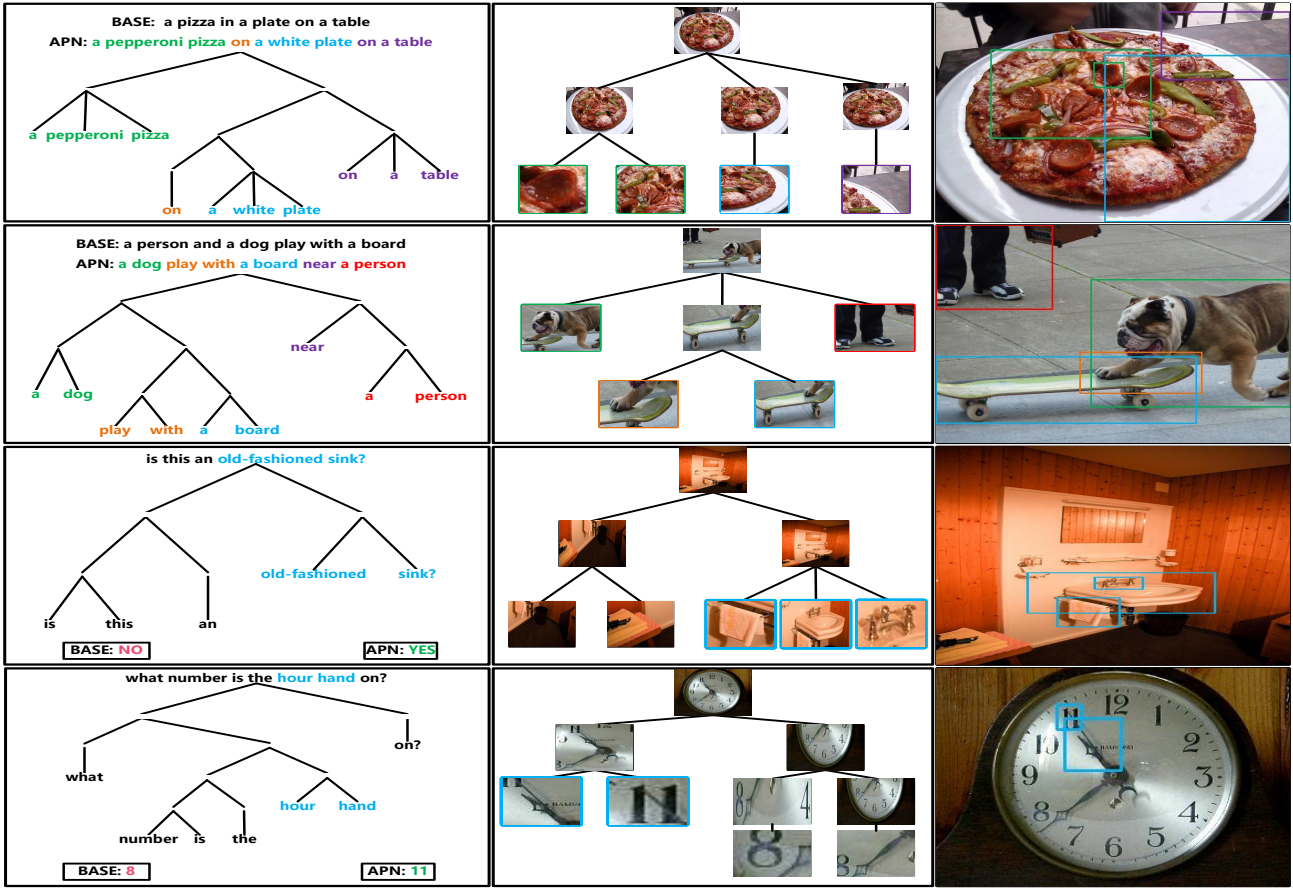}
   \caption{The parsed trees. The first and last two rows are examples for captioning and VQA, respectively. For captioning, the probabilities of both $N_V$ and $N_L$ layers in Figure~\ref{fig:fig_capach} are used. For VQA, the probabilities of $N_C$ layers in Figure~\ref{fig:fig_vqaach} are used.
   The same colors between language and visual trees show the alignments. The red/green color in VQA cases indexes the correct/wrong answers. }
\label{fig:fig_vis_ca}
\vspace{-0.2in}
\end{figure*}

\begin{table}[t]
\begin{center}
\caption{Learderboard of various methods using single model on the MS-COCO online test server.}
\label{table:tab_cap_on}
\scalebox{0.65}{
\begin{tabular}{l c c c c c c c c}
		\hline
		\multirow{2}*{Models}   & \multicolumn{2}{c}{B@4} & \multicolumn{2}{c}{M} & \multicolumn{2}{c}{R} &  \multicolumn{2}{c}{C} \\ \cmidrule(r){2-3} \cmidrule(r){4-5} \cmidrule(r){6-7} \cmidrule(r){8-9} 
		   & c5 & c40 & c5 & c40 & c5 & c40 & c5 & c40\\
		    \hline
           SCST~\cite{rennie2017self}  & $35.2$  & $64.5$  & $27.0$  & $35.5$  & $56.3$  & $70.7$  & $114.7$  & $116.0$    \\ 
           LSTM-A~\cite{yao2017boosting}& $35.6$  & $65.2$  & $27.0$  & $35.4$  & $56.4$  & $70.5$  & $116.0$  & $118.0$   \\ 
           Up-Down~\cite{anderson2018bottom}  & $36.9$  & $68.5$  & $27.6$  & $36.7$  & $57.1$  & $72.4$  & $117.9$  & $120.5$ \\ 
           RFNet\cite{jiang2018recurrent} & $38.0$  & $69.2$  & $28.2$  & $37.2$  & $58.2$  & $73.1$  & $122.9$  & $125.1$ \\ 
           SGAE~\cite{yang2019auto} & $37.8$  & $68.7$  & $28.1$  & $37.0$  & $58.2$  & $73.1$  & $122.7$  & $125.5$ \\
           CNM~\cite{yang2019learning} & $38.4$  & $69.3$  & $28.2$  & $37.2$  & $58.4$  & $73.4$  & $123.8$  & $126.0$ \\
           ETA~\cite{li2019entangled} & $38.9$ & $\mathbf{70.2}$ & $28.6$ & $\mathbf{38.0}$ & $58.6$  & $73.9$  & $122.1$  & $124.4$ \\ 
           AoANet~\cite{huang2019attention}  & $37.3$ & $68.1$ & $28.3$ & $37.2$ & $57.9$ & $72.8$ & $124.0$ & $126.2$ \\
           \hline
           APN    & $\mathbf{38.9}$ & $\mathbf{70.2}$ & $\mathbf{28.8}$ & $\mathbf{38.0}$ & $\mathbf{58.7}$ & $\mathbf{73.7}$ & $\mathbf{126.3}$ & $\mathbf{127.6}$\\ \hline
\end{tabular}
}
\end{center}
\vspace{-0.3in}
\end{table}
\noindent\textbf{Qualitative Results}.
We visualize the language and image trees parsed by Algorithm~\ref{alg1} in the first two rows of Figure~\ref{fig:fig_vis_ca}. To parse trees, we use the $M$ of all $N_L$ and $N_V$ self-attention layers in Figure~\ref{fig:fig_capach}. The left and middle parts show the trees of the generated caption and the given image, respectively. Intuitively, our APN generates captions in a phrase-by-phrase manner instead of a word-by-word manner because our model considers previous words when  generating new words at the phrase level as shown in Eq.~\eqref{equ:bernoulli_distribution}. For example, in the first row of Figure~\ref{fig:fig_vis_ca}, when generating the word “pizza”, our captioner first considers whether it is in the same cluster with “a pepperoni”. Furthermore, the PGM probability matrix $\bm{M}$ calculated by Eq.~\eqref{equ:mij} contains the local context of clusters - phrases, and the phrase-level local information is incorporated when generating the captions. In this way, more details will be described, compared to BASE that only describes ``a pizza''. Besides, the second case demonstrates that APN alleviates the bias. Since ``person'' and ``board'' have high co-occurrence in the training set, Base easily learns such bias to generate ``person-play-board'', while APN learns the significant local context ``dog-play-board'' and then generates the right description. Also, by comparing language and visual trees, we can see that they have similar structures, \eg, in the ``pizza case'', the leftmost leaves of both trees focus on ``a pepperoni pizza'' and the rightmost leaves focus on ``a table''. This observation suggests that the hidden structures are transferred from the language domain into the vision domain. 

\noindent\textbf{Comparisons to State-of-the-art methods}.
We compare our APN with various state-of-the-art models trained by the RL-based reward (Eq.~\eqref{equ:equ_rlloss}) which are GCN-LSTM~\cite{yao2018exploring},  SGAE~\cite{yang2019auto}, CNM~\cite{yang2019learning}, HIP~\cite{yao2019hierarchy}, AoA~\cite{huang2019attention}, ORT~\cite{herdade2019image}, ETA~\cite{li2019entangled} and $\mathcal{M}^2$ Transformer~\cite{cornia2020meshed}. Specifically, GCN-LSTM, SGAE, CNM, and HIP incorporate additional graph annotations into the model, \eg, GCN-LSTM exploits the objects' pairwise relationships acquired from a pre-trained relation classifier. For AoA, ORT, ETA and $\mathcal{M}^2$ Transformer, all of them exploit Transformer as the backbone. Note that we only compare with the captioners that have a similar scale as ours and do not compare with certain large-scale captioners like OSCAR~\cite{li2020oscar}.
 
Table~\ref{table:tab_cap} reports the results of various captioners on the Karpathy test split. We can see that our APN achieves a higher CIDEr-D score than the other state-of-the-art models. In particular, compared with the self-attention based models, \eg, ORT and AOA, APN transfers the dense graph structure into a sparse one, which encourages the captioner to exploit more crucial local contexts for captioning. Also, compared with the captioners that use graph data generated by other pre-trained parsers, \eg, SGAE or HIP, our APN suffers less from the distribution shifts. We also report our single model's performance on the MS COCO online test server in Table~\ref{table:tab_cap_on}, which demonstrates that our model achieves better performances than the other ones in terms of most metrics.  We do not compare with~\cite{pan2020x} because X-LAN needs larger batch size and more GPUs. Furthermore, we plan to extend our PGM framework to other state-of-the-art transformers, \eg, AoANet~\cite{huang2019attention}, $\mathcal{M}^2$ Transformer~\cite{cornia2020meshed} and X-LAN~\cite{pan2020x} in the future.

\begin{figure}[t]
\centering
\includegraphics[height=7.6cm, width=1\linewidth]{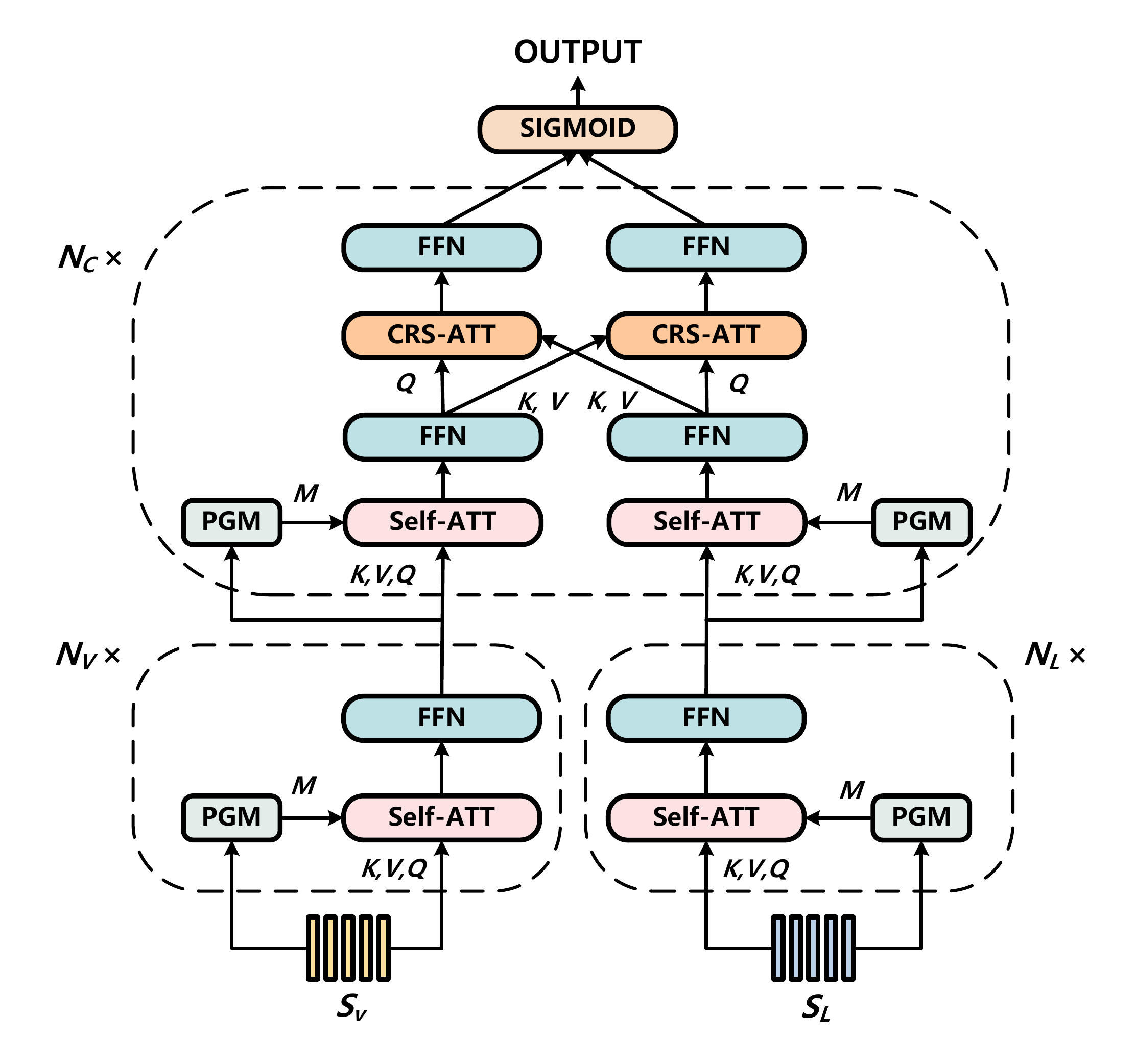}
   \caption{An overview of our APN for VQA. The FFN, CRS-ATT and Self-ATT denote the feed-forward network, cross-attention network and self-attention network, respectively. 
   }
\label{fig:fig_vqaach}
\vspace{-0.2in}
\end{figure}

\section{Visual Question Answering}
\subsection{Architecture}
The APN architecture for VQA is illustrated in Figure~\ref{fig:fig_vqaach}, where $S_V$ and $S_L$ denote the embeddings of the images and the questions. Note that the $S_V$ is linearized by sorting the RoIs from top-left to bottom-right. In experiments, we set $N_{V}=1$, $N_{L}=6$, and $N_{C}=6$. In particular, $N_{V}$ is set to 1 because the input visual features are extracted from the last few layers of Faster-RCNN~\cite{ren2015faster,anderson2018bottom}, which have already captured abundant high-level knowledge. The outputs of the cross-attention include two parts, which are added and then projected to a 3,129-dimensional vector, where 3,129 is the number of the most frequent answers. We follow~\cite{teney2018tips} to train our model by the binary cross-entropy loss.

\subsection{Dataset, Settings and Metrics}
\noindent\textbf{VQA-v2 Dataset}~\cite{antol2015vqa} includes images from the MS-COCO dataset, with 3 questions for each image and 10 answers per question. It has 80k training images and 40k validation images available offline. The online evaluation provides test-dev and test-std splits, each of which has 80k images. For offline evaluation, the validation images are split into two sets, typically for validation and test, separately.

\noindent\textbf{Settings}.
We set the dimensionality of the image feature, the question embedding and the multi-modal embedding to $2,048$, $512$ and $1,024$, respectively. In the ablation studies, we set the hidden size of the multi-head attention to 512 and only train our model on the training set of the VQA2.0 dataset. We multiply all heads' attention matrix by the same $M$ (Eq.~\eqref{equ:multi-head}). When compared with the other state-of-the-art models, we follow the conventions to set the hidden size of the multi-head attention to 1024 and also exploit the question-answer pairs in Visual Genome~\cite{krishna2017visual} to train our model. We apply the Adam\cite{kingma2014adam} optimizer to train our model and follow MCAN~\cite{yu2019deep} to set the learning rate to $\textbf{min}(2.5te^{-5},1e^{-4})$, where $t$ is the training epoch and after 10 epochs, the learning rate decays by 0.2 every 2 epochs. The batch size is 64 and the training epoch is 13. 
\subsection{Results}
\noindent\textbf{Ablation Studies}. 
We design the similar baselines \textbf{BASE}, \textbf{PGM}, and \textbf{APN} as in Section~\ref{sec:ImgCap} to test the effectiveness of our two key implementations: the PGM module and the hierarchical constraint. We report the accuracies of different question types on offline local validation split to compare the performances of these models, which are reported in Table~\ref{table:tab_vqa_ab}. From this table, we can see that our APN achieves the highest accuracies on all question types. Hence, the extracted sparse and hierarchical structures are useful for VQA models to get more correct answers.

\begin{table}[t]
\begin{center}
\caption{VQA-v2 Val accuracy scores of ablative baselines.}
\label{table:tab_vqa_ab}
\scalebox{1}{
\begin{tabular}{l c c c c}
		\hline
		    Models & Yes/No & Number & Other & Overall  \\ \hline
		    BASE  & $83.30$  & $47.95$ & $57.30$ & $65.84$\\ 
		  PGM  & $83.90$  & $48.63$ & $57.92$ & $66.51$\\ 
		   APN  & $\mathbf{84.99}$  & $\mathbf{49.71}$ & $\mathbf{58.66}$ & $\mathbf{67.38}$\\ 
		   \hline
\end{tabular}
}
\end{center}
\vspace{-0.2in}
\end{table}

\begin{table}[t]
\begin{center}
\caption{VQA-v2 test-dev and test-std accuracy of various models.}
\label{table:tab_vqa_st}
\scalebox{0.8}{
\begin{tabular}{l c c c c c}
		\hline

		    \multirow{2}*{Models}  & \multicolumn{4}{c}{Test-dev}  & Test-std\\ \cmidrule(r){2-5} \cmidrule(r){6-6} 
		   & Yes/No & Number & Other & Overall  & Overall \\
		    \hline
		  DCN~\cite{nguyen2018improved}  & $83.51$  & $46.61$ & $57.26$ &$66.87$& $66.97$ \\ 
		  VCTREE~\cite{tang2019learning}  & $84.28$  & $47.78$ & $59.11$ &$68.19$&  $68.49$\\ 
		  BAN~\cite{kim2018bilinear}  & $85.42$  & $54.04$ & $60.52$ &$70.04$& $70.35$ \\ 
		  DFAF~\cite{gao2019dynamic}  & $86.09$  & $53.32$ & $60.49$ &$70.22$&  $70.34$\\ 
		   MCAN~\cite{yu2019deep}  & $85.82$  & $53.26$ & $60.72$ &$70.63$ &$70.90$\\ 
		   TRRNet~\cite{yang2020trrnet}  & $87.27$  & $51.89$ & $61.02$ &$70.80$ &$71.20$\\ 
		   \hline 
		   APN  & $\bm{87.44}$  & $\bm{52.68}$ & $\bm{61.18}$ &$\bm{71.14}$ &$\bm{71.33}$\\ 
		   \hline
           
\end{tabular}
}
\end{center}
\vspace{-0.3in}
\end{table}
\noindent\textbf{Qualitative Results.}
We use the $M$ from all $N_c=6$ self-attention layers of the decoder in Figure~\ref{fig:fig_vqaach} to parse the trees. Two examples are illustrated in the last two rows of Figure~\ref{fig:fig_vis_ca} to demonstrate how the local contexts help the model correctly answer the questions. For example, we can find that the local context, \eg, "old-fashioned sink", "hour hand" is gathered in the same cluster for both language and visual trees. In this way, our APN can accurately understand the questions and then attend to the right regions of an image to get the correct answers. Without recognizing the local context, BASE model may neglect the essential adjectives and thus fall into the wrong answers. For example, in the last row, the answer of BASE is "8" to the question, which is confused by the number under the minute hand.

\noindent\textbf{Comparisons to State-of-the-art methods}
We compare our APN with certain state-of-the-art models that exploit graph priors or are designed based on the Transformer. Their performances are reported in Table~\ref{table:tab_vqa_st}. Note that we do not compare our method with the large-scale pre-training models like ERNIE-VIL~\cite{yu2020ernie}. From this table, we can see that our APN achieves higher accuracies than the others. For example, compared with VCTREE~\cite{tang2019learning} which also incorporates sparse tree priors or TRRNet~\cite{yang2020trrnet} which is built based on Transformer, our APN gets the highest accuracies. These comparisons suggest that APN can utilize the sparse tree more effectively than VCTREE, and using the sparse and hierarchical structure has advantages in solving VQA task over the Transformer's fully connected structure.

\section{Conclusion}
In this paper, we impose a Probabilistic Graphical Model (PGM) on the self-attention layers of a Transformer to incorporate the sparse assumption into the original fully connected one. Then, trivial global dependencies can be avoided, and critical local context can be discovered and exploited. Furthermore, we stacked the constrained self-attention layers and imposed hierarchical constraints on them by which a tree can be implicitly parsed. In this way, the model can unsupervisedly parse trees during the end-to-end training. A tree parsing algorithm was also provided, which exploits the calculated PGM probabilities to extract the hidden trees. Thus, we can figure out the hidden structure of each sample. We proposed two different APNs on Image Captioning and Visual Question Answering, and the results demonstrate that APN can improve both tasks compared with self-attention based Transformers.

\noindent\textbf{Acknowledgements.} 
This work is partially supported by NTU TIER2 and Monash FIT Start-up Grant.

{\small
\bibliographystyle{ieee_fullname}
\bibliography{egpaper_final}
}

\end{document}